\documentclass{article}

    \usepackage[final]{neurips_2024}

\usepackage[utf8]{inputenc} 
\usepackage[T1]{fontenc}    
\usepackage{hyperref}       
\usepackage{url}            
\usepackage{booktabs}       
\usepackage{amsfonts}       
\usepackage{nicefrac}       
\usepackage{microtype}      
\usepackage{xcolor}         
\usepackage{hyperref}
\usepackage{url}
\usepackage[pdftex]{graphicx}
\usepackage{xcolor}         
\usepackage{amsmath}
\usepackage{amssymb}
\usepackage{amsthm}
\newtheorem{theorem}{Theorem}

\newtheorem{proposition}{Proposition}
\usepackage{algorithmic}
\usepackage{algorithm}
\usepackage{subfig}
\usepackage{wrapfig}
\usepackage{multirow}
\usepackage{multicol}
\usepackage{threeparttable}
\usepackage{array}
\usepackage{hyperref}       
\usepackage{url}            
\usepackage{booktabs}       
\usepackage{amsfonts}       
\usepackage{nicefrac}       
\usepackage{microtype}      

\title{Improving Molecular Graph Generation with \\ Flow Matching and Optimal Transport}

\author{
{Xiaoyang Hou}$^{1,2}$ \thanks{Contributed equally: Author order is randomized and can be adjusted as needed for individual purposes.}
\quad Tian Zhu$^{1,2}$ \footnotemark[1]
\quad {Milong Ren}$^{1,2}$
\quad {Dongbo Bu}$^{1,2,3}$ \\
\textbf{Xin Gao}$^{6}$ 
\quad \textbf{Chunming Zhang}$^{1,4,5}$
\quad \textbf{Shiwei Sun}$^{1,2,4}$
\thanks{Correspondence should be addressed to Shiwei Sun (dwsun@ict.ac.cn)}
\\
1. Key Laboratory of Intelligent Information Processing, Institute of Computing Technology. \\ 
2. School of Computer Science and Technology, University of Chinese Academy of Sciences. \\
3. Central China Research Institute for Artificial Intelligence Technologies. \\
4. Western Institute of Computing Technology.  \\
5. Phil Rivers Technology. \\
6. King Abdullah University of Science and Technology. 
}

\begin{document}

\maketitle

\begin{abstract}
Generating molecular graphs is crucial in drug design and discovery but remains challenging due to the complex interdependencies between nodes and edges. While diffusion models have demonstrated their potentiality in molecular graph design, they often suffer from unstable training and inefficient sampling. To enhance generation performance and training stability, we propose GGFlow, a discrete flow matching generative model incorporating optimal transport for molecular graphs and it incorporates an edge-augmented graph transformer to enable the direct communications among chemical bounds. Additionally, GGFlow introduces a novel goal-guided generation framework to control the generative trajectory of our model, aiming to design novel molecular structures with the desired properties. GGFlow demonstrates superior performance on both unconditional and conditional molecule generation tasks, outperforming existing baselines and underscoring its effectiveness and potential for wider application.
\end{abstract}

\section{Introduction}

De novo molecular design is a fundamental but challenging task in drug discovery and design. While the searching space of the molecular graph is extremely tremendous, as large as $10^{33}$ \citep{polishchuk2013estimation}. Machine learning methods have been introduced to generate molecular graphs due to the large amount of data in the field. These models are typically categorized into autoregressive and one-shot types. Autoregressive models, such as GraphDF \citep{luo2021graphdf}, generate graphs sequentially, often overlooking the interdependencies among all graph components. In contrast, one-shot methods generate entire graphs in a single step, more effectively capturing the joint distribution \citep{kong2022molecule}.

Diffusion models have shown great promise and achieved significant performance in various domains \citep{ho2020denoising, song2020score, ho2022video}. In the context of molecular graph generation, diffusion models have been adopted to enhance generative capacity. EDP-GNN and GDSS are among the first to utilize diffusion models for graph generation, adding continuous Gaussian noise to adjacency matrices and node types, which may lead to invalid molecular graph structures \citep{niu2020permutation, jo2022score}. Due to the inherent sparsity and discreteness of molecular graph structures, GSDM enhances model fidelity by introducing Gaussian noise within a continuous spectrum space of the graph, and DiGress and CDGS apply discrete diffusion models for graphs \citep{luo2023fast, vignac2022digress, austin2021structured, haefeli2022diffusion, huang2023conditional}.

Despite their potential, diffusion models often face challenges with unstable training and inefficient sampling. Flow matching generative models offer a more stable and efficient alternative by transforming the generative process from stochastic differential equations (SDEs) to ordinary differential equations (ODEs), enhancing generative efficiency \citep{lipman2022flow, song2024equivariant, yim2023fast}. Additionally, the use of optimal transport straightens the marginal probability path, reducing training variance and speeding up sampling \citep{bose2023se, tong2023improving, klein2024equivariant, pooladian2023multisample}. However, the application of OT in graph-based systems is often hampered by significant computational demands, primarily due to the complexity of the OT metric \citep{chen2020graph, petric2019got}.

In this paper, we introduce GGFlow, a novel generative model that leverages discrete flow matching techniques with optimal transport to improve sampling efficiency and training stability in molecular graph generation. GGFlow incorporates an edge-augmented graph transformer to model direct chemical bond relations, benefiting chemical bond generation tasks. The model preserves graph sparsity and permutation invariance, essential for valid molecular graph generation. Additionally, GGFlow employs a goal-guided framework using reinforcement learning for molecule design with target properties. GGFlow achieves state-of-the-art results in both unconditional and conditional molecule generation tasks and surpasses existing methods with fewer inference steps. Its effectiveness in conditional generation tasks underscores the practical impact of our approach.

Our contribution can be summarized as:
\begin{itemize}
    \item GGFlow introduces the first discrete flow matching generative model with optimal transport for molecular graph data, improving sampling efficiency and training stability. It also incorporates an edge-augmented graph transformer to enhance generation tasks.
    \item GGFlow proposes a novel guidance framework using reinforcement learning to control probability flow during molecular graph generation, targeting specific properties.
    \item GGFlow demonstrates state-of-the-art performance in various unconditional and conditional molecular graph generation tasks, consistently outperforming existing methods across diverse graph types and complexities.
\end{itemize}

\section{Related Work}

\subsection{Flow Matching and Diffusion Models}
Diffusion models have gained widespread popularity in various fields, including computer vision, natural language processing, and biological sciences, demonstrating notable success in generative tasks \citep{ho2020denoising, song2020score, watson2023novo, ingraham2023illuminating, liu2024predicting, ren2024carbonnovo,zhu2024antibody}. However, these models often suffer from inefficiencies in sampling due to the complexity of their underlying diffusion processes and the convergence properties of the generative process.

\begin{figure}[!htp]
    \centering
    \includegraphics[scale=0.25]{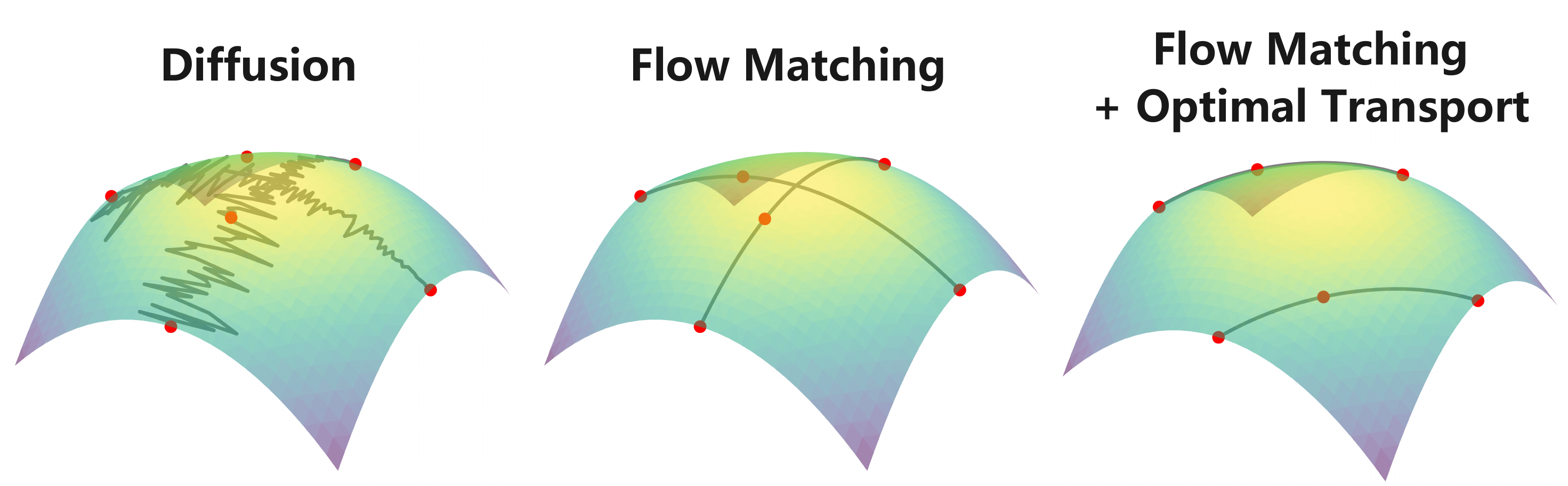}
    \caption{Illustration of generative trajectories using different methods. The generative trajectories are learned by the diffusion model (left), flow matching model (center), and flow matching model with optimal transport (right).}
    \label{fig:overview}
\end{figure}

Flow matching generative models have emerged as a more efficient and stable alternative (details in Appendix \ref{app:flow}), improving sampling by straightening the generative probability path \citep{lipman2022flow, song2024equivariant, campbell2024generative}. Some approaches further enhance performance by incorporating optimal transport. The generative processes of these models are summarized in Figure \ref{fig:overview}.


Previous works \citep{campbell2024generative, gat2024discrete} extended flow matching to discrete spaces, while \citet{eijkelboom2024variational} applied variational flow matching to graphs, but without adequately addressing key graph-specific properties such as adjacency matrix sparsity. GGFlow tackles these challenges by introducing a discrete flow matching model with optimal transport tailored for graph data. Furthermore, we propose a novel framework for guiding the generative process, enhancing its practical applicability.

\subsection{Molecular Graph Generative Models}
Molecular Graph generative models are typically categorized into two main types: autoregressive and one-shot models. Autoregressive models, such as generative adversarial networks \citep{wang2018graphgan}, recurrent neural networks \citep{you2018graphrnn}, variational autoencoders \citep{jin2018junction}, normalizing flows \citep{shi2019graphaf, luo2021graphdf} and diffusion model \citep{kong2023autoregressive}, generate graphs sequentially. While effective, these models are often computationally expensive and fail to account for permutation invariance, a crucial property for graph data, resulting in potential inefficiencies. In contrast, one-shot models aim to capture the distribution of all molecular graph components simultaneously \citep{de2018molgan, ma2018constrained, zang2020moflow}, better reflecting the inherent interactions within molecular graphs. Despite the advantages, diffusion-based one-shot models \citep{niu2020permutation, jo2022score, vignac2022digress, chen2023efficient, bergmeister2023efficient, luo2023fast, haefeli2022diffusion, yan2023swingnn, jang2023simple, madeira2024generative, bergmeisterefficient, chen2023efficient, minello2024graph,zhao2024pard,xu2024discrete} show promising results in downstream tasks but remain limited by sampling efficiency. GGFlow addresses these limitations by employing a discrete flow-matching generative model, achieving superior generative performance with fewer sampling steps.

\section{Methods}
In this section, we present our methodology, GGFlow. Section \ref{sec:fm} outlines the discrete flow matching method for molecular graph generation. Section \ref{sec:ot} covers optimal transport for graph flow matching. Section \ref{sec:evo} introduces GraphEvo, our neural network for graph generation. Section \ref{sec:theo} examines the permutation properties of GGFlow, and Section \ref{sec:rl} discusses goal-guided molecule generation using reinforcement learning.

\subsection{Discrete Flow Matching for Molecular Graph Generation}
\label{sec:fm}

A molecular graph $G = (V, E)$, where $V$ and $E$ denote the sets of nodes and edges, has a distribution denoted by $p(G) = (p^V(V), p^E(E))$. The attribute spaces for nodes and edges are $\mathcal{V}$ and $\mathcal{E}$, with cardinalities $n$ and $m$, respectively. The attributes of node $i$ and edge $ij$  are denoted by $v_i \in \mathcal{V}$ and $e_{ij} \in \mathcal{E}$, so the node and edge probability mass functions (PMF) are $p^V(v_i=a)$ and $p^E(e_{ij}=b)$ where $a\in \{1,\dots, n\}$ and $b \in \{1,\dots, m\}$. The node and edge encodings in the graph are given by matrices $\mathbf{V} \in \mathbb{R}^{a \times n}$ and $\mathbf{E} \in \mathbb{R}^{a \times a \times m}$, respectively. We denote the transpose of matrix $\mathbf{A}$ as $\mathbf{A}^*$ and $\mathbf{A}^t$ represents the state of matrix $\mathbf{A}$ at time $t$. We use discrete flow matching to model the molecular graph generation process.

\textbf{Source and target distribution}\quad
GGFlow aims to transform prior distribution $G^0\sim p_{\rm ref}$ to target data distribution $G^1 \sim p_{\rm data}$. The training data $(G^0, G^1)$ are sampled from a joint distribution $\pi(G^0, G^1)$, satisfying the marginals constraints $p_{\rm ref}=\sum_{G^1}\pi(G^0, G^1), p_{\rm data}=\sum_{G^0}\pi(G^0, G^1)$. In the simplest case, the joint distribution $\pi(G^0, G^1)$ is modeled as the independent coupling, i.e. $\pi(G^0,G^1)=p_{\rm ref}\cdot p_{\rm data}$. 

To account for graph sparsity, the prior distribution $p_{\rm ref}= (p^V_{\rm ref}, p^E_{\rm ref})$ is designed to approximate the true data distribution closely. To ensure the permutation invariance of the model, the priors are structured as products of single distributions for all nodes and edges: $\prod_i v_i \times \prod_{ij} e_{ij}$ \citep{vignac2022digress}. Further details on the prior can be found in Appendix \ref{app:prior}.

\textbf{Probability path}\quad
We define a probability path $p_t(G^t)$ that interpolates between source distribution $p_{\rm ref}$ and target distribution $p_{\rm data}$ i.e. $p_0 = p_{\rm ref}$ and $p_1 = p_{\rm data}$. The marginal probability path is given by: 
\begin{align}
    \label{eq:path}
    p_t(G^t) &= \sum_{(G^0,G^1)\sim\pi}p_t(G^t|G^0,G^1)\pi(G^0,G^1), 
\end{align}
where 
\begin{align*}
    p_{t}(G^t|G^0,G^1) &= {\rm Cat}\Big (t\delta\{G^1,G\}+(1-t)p_{\rm ref}\Big ) \\
    &= {\rm Cat}\Big (t\delta\{V^1,V\}+(1-t)p^V_{\rm ref}, t\delta\{E^1,E\}+(1-t)p^E_{\rm ref}\Big ),
\end{align*}
\( \delta \) is the Kronecker delta, indicating equality of the indices, and ${\rm Cat}(p)$ denotes a Categorical distribution with probabilities $p$. Given the sparsity of both the prior and data distributions, we can infer that the intermediate distribution is similarly sparse, aiding model training.

We define a probability velocity field $u_t(G, G^t) = (u_t^V(V, V^t), u_t^E(E, E^t))$ for GGFlow, which generates the probability path from Equation \ref{eq:path}. The probability velocity field $u_t(G,G^t)$ is derived from the conditional probability velocity field $u_t(G, G^t|G^0, G^1)$, and can be expressed as: 
\begin{align}
    u_t(G,G^t) &= \sum_{(G^0,G^1)\sim \pi}u_t(G, G^t|G^0, G^1)p_t(G^0,G^1|G^t), \\
    p_t(G^0,G^1|G^t) &= p_{1|t}(G^1|G^t,G^0)\frac{p_{t}(G^t|G^0,G^1)\pi(G^0,G^1)}{\sum_{G^0,G^1}p_t(G^t|G^0,G^1)\pi(G^0,G^1)}.
\end{align}

GGFlow chooses the conditional marginal probability $u_t(G, G^t|G^0, G^1)$ as: 
\begin{equation}
    u_t(G, G^t|G^0,G^1) = \frac{1}{\mathbf{Z}_t(1-t)p_{\rm ref}}\delta\{G,G^1\}(1-\delta\{G^t,G^1\}), G_t \neq G,
\end{equation}
where ${\rm ReLU(a) = max(a,0)}$ and \( \mathbf{Z}_t = |\{G^t: p_{t}(G^t|G^0,G^1)>0\}| \). More details about the conditional vector field are provided in Appendix \ref{app:p1}.

\textbf{Training objective}\quad
Given the intractability of the posterior distribution $p_{1|t}(G^1|G^t,G^0)$, we approximate it as $\hat{p}_{1|t}(G^1|G^t,G^0)$ using neural network, as detailed in Section \ref{sec:evo}. The training objective is formulated as: 
\begin{align}
    \label{loss}
    \mathcal{L} = \mathbb{E}_{p_{\rm data}(G^1)\mathcal{U}(t;0,1)\pi(G^0,G^1)p_{t}(G^t|G^0,G^1)}[\log \hat{p}_{1|t}(G^1|G^t,G^0)],
\end{align}
where $\mathcal{U}(t;0,1)$ is a uniform distribution on $[0,1]$.

\textbf{Sampling Procedure}\quad
In the absence of the data distribution $G^1$ during sampling, we reparameterize the conditional probability $p_t(G^0,G^1|G^t)$ as:
\begin{align*}
    p_t(G^0,G^1|G^t) &= p_{1|t}(G^1|G^t,G^0)\frac{p_{t}(G^t|G^0)p(G^0)}{\sum_{G^0}p_t(G^t|G^0)p(G^0)}. \\
    p_{t}(G^t|G^0) &= {\rm Cat}\Big (t\delta\{V^1,V\}+(1-t)p_V^{\rm ref}, t\delta\{E^1,E\}+(1-t)p_E^{\rm ref}\Big )
\end{align*}
And we can simplify the generative process $p_{t+\Delta t|t}(G^{t+\Delta t} | G^t, G^0)$ without the calculation of the full expectation over conditional vector field $\hat{u}_t(G,G^t|G^0,G^1)$: 
\begin{align}
    p_{t+\Delta t|t}(G^{t+\Delta t} | G^t, G^0) &= \mathbb{E}_{\hat{p}_{1|t}(G^1|G^t,G^0)}[\delta(G^t,G^{t+\Delta t})+u_{t}(G^t,G^{t+\Delta t}|G^0,G^1) \Delta t] \notag \\
    &=\sum_{G^1}p_{t+\Delta t|t}(G^{t+\Delta t}|G^1,G^t,G^0)\hat{p}_{1|t}(G^1|G^t,G^0).
\end{align}

We first sample the $\hat{G}^1$ using the approximate distribution $\hat{p}_{1|t}(G^1|G^t,G^0)$ and then sample the next state $G^{t+\Delta t}$ using sampled $\hat{G}^1$. The sampling procedure $p_{t+\Delta t|t}(G^{t+\Delta t}|G^1,G^t,G^0)$ can thus be formulated as:
\begin{equation}
    G^{t+\Delta t} \sim \delta\{\cdot,G^t\} + {u}_t(\cdot,G^t|G^0,\hat{G}^1)\Delta t.  \notag
\end{equation}
Further details on the sampling and training procedures are provided in Algorithms \ref{alg:sample} and \ref{alg:train}.
\begin{algorithm}[!htbp]
    \small
    \caption{Sampling Procedure of GGFlow}\label{alg:sample}
    \begin{algorithmic}[1]
        \REQUIRE $t=0, G^0\sim (p_V^{\rm ref}, p_E^{\rm ref}), u_t(G,G^t|G^0, G^1), N_{\rm steps}$
        \STATE $\Delta t = 1/N_{\rm steps}$
        \FOR{$n \in \{0,\dots,N_{\mathrm{steps}}-1\}$}
        \STATE $\hat{p}_{1|t}(G^1|G^0,G^t) = {\rm GraphEvo}(G^t,G^0,t)$
        \STATE $\hat{G}^1 \sim \hat{p}_{1|t}(\cdot|G^0,G^t)$
        \STATE // Sampling from the conditional velocity field
        \STATE $G^{t+\Delta t}\sim \delta\{\cdot,G^t\} + {u}_t(\cdot,G^t|G^0, \hat{G}^1)\Delta t $
        \STATE $t = t+\Delta t$
        \ENDFOR
        \RETURN $G^1 = (V^1, E^1)$
    \end{algorithmic}
\end{algorithm}

\subsection{Optimal transport for graph flow matching}
\label{sec:ot}
Optimal transport (OT) has been effectively applied to flow matching generative models in continuous variable spaces, to improve generative performance \citep{tong2023improving, bose2023se, song2024equivariant}. To generalize this for graphs, we extend the joint distribution $\pi(G^0, G^1)$ from independent coupling to the 2-Wasserstein OT map $\phi^*$, which minimizes the 2-Wasserstein distance between $p_{\rm ref}$ and $p_{\rm data}$. To optimize the computational efficiency of OT and preserve permutation invariance, we define the distance via the Hamming distance $H(G^1, G^0)$ \citep{bookstein2002generalized}:
\begin{align}
    \label{eq:ot}
    \phi^*(p_0,p_1) = \arg\inf_{\phi \in \Phi} \int_{\mathbb{R}^d\times \mathbb{R}^d}H(G^0,G^1)\mathrm{d} \phi(G^0,G^1), 
\end{align}
where 
\begin{equation}
    H(G^0,G^1) = \sum_{i} \delta(v_i^0, v_i^1) + \lambda \sum_{i,j} \delta(e_{ij}^0, e_{ij}^1).
\end{equation}
Here \( \Phi \) represents the set of all joint probability measures on \( \mathbb{R}^d \times \mathbb{R}^d \) that are consistent with the marginal distributions \( p_0 \) and \( p_1 \), where  \( G^K = (V^K=\{v_i^K\}, E^K=\{e_{ij}^K\}_{ij}) \), \( K=0, 1 \).

The practical application of OT to large datasets is computationally intensive, often requiring cubic time complexity and quadratic memory \citep{tong2020trajectorynet, villani2009optimal}. To address these challenges, we use a minibatch approximation of OT \citep{fatras2021minibatch}.

\subsection{GraphEvo: Edge-augmented Graph Transformer}
\label{sec:evo}

Our neural network, GraphEvo, predicts the posterior distribution \( p_{1|t}(G^1 | G^t, G^0) \) using the intermediate graph \( G^t \). In graph-structured data, edge and structural information are as critical as node attributes, and incorporating edge relations enhances chemical bond generation tasks \citep{hussain2024triplet, hou2024gtam, jumper2021highly}. To capture these relations, GraphEvo extends the graph transformer by introducing a triangle attention mechanism for edge updates, along with additional graph features \( y \), such as cycles and the number of connected components \citep{vignac2022digress}. 
This enables GraphEvo to efficiently and accurately capture the joint distribution of all graph components. The key self-attention mechanisms are outlined in Algorithm \ref{alg:evo}, where node, edge, and graph features are represented as \( \mathbf{X} \in \mathbb{R}^{bs \times n \times dx} \), \( \mathbf{E} \in \mathbb{R}^{bs \times n \times dx} \), and \( \mathbf{y} \in \mathbb{R}^{bs \times n \times dy} \), where \( bs \) denotes batch size, \( n \) is the number of nodes, and \( dx \) and \( dy \) are the feature dimensions for node and global features, respectively. Further details are provided in Appendix \ref{app:graphevo}.

\begin{algorithm}[!htbp]
    \caption{Self-attention Mechanism in GraphEvo}\label{alg:evo}
    \begin{algorithmic}[1]
        \REQUIRE $\mathbf{X}\in \mathbb{R}^{bs\times n\times dx},\mathbf{E}\in \mathbb{R}^{bs\times n\times dx},\mathbf{y}\in \mathbb{R}^{bs\times n\times dy}$
        \STATE $\mathbf{Q},\mathbf{K},\mathbf{V} \leftarrow {\rm Linear}(\mathbf{X})$ 
        \STATE $\mathbf{Y} \leftarrow \frac{\mathbf{Q}\times \mathbf{K}}{\sqrt{d_\mathbf{Y}}}$\quad // Calculation attention score for node embedding 
        \STATE $\mathbf{Y} \leftarrow {\rm FiLM}(\mathbf{Y}, \mathbf{E}) $ \quad // Incorporate edge features to self-attention scores
        \STATE $\mathbf{E} \leftarrow \mathbf{Y}$ 
        \STATE $\mathbf{Q_e}, \mathbf{K_e}, \mathbf{V_e}, \mathbf{b}, \mathbf{g} \leftarrow {\rm Linear}(\mathbf{E})$
        \STATE $\mathbf{Y_e} \leftarrow \frac{\mathbf{Q_e}\times \mathbf{K_e}}{\sqrt{d_\mathbf{Y_e}}} + \mathbf{b}$ \quad // Calculation triangle attention score for edge embedding
        \STATE $\mathbf{E} \leftarrow \mathbf{Y_e}*\mathbf{V_e}*{\rm sigmoid}(\mathbf{g})$ 
        \STATE $\mathbf{E} \leftarrow {\rm Linear}\Big ({\rm FiLM}(\mathbf{E}, \mathbf{y})\Big ) $ \quad // Incorporate global structural features to edge embedding
        \STATE $\mathbf{X} \leftarrow \mathbf{Y} * \mathbf{V}$ 
        \STATE $\mathbf{X} \leftarrow {\rm Linear}\Big ( {\rm FiLM}(\mathbf{X}, \mathbf{y}) \Big )$ \quad // Incorporate global structural features to node embedding
        \STATE $\mathbf{y} \leftarrow {\rm Linear}\Big ({\rm Linear}(\mathbf{y}) + {\rm PNA}(\mathbf{X}) + {\rm PNA}(\mathbf{E})\Big )$
        \RETURN $\mathbf{X},\mathbf{E},\mathbf{y}$
    \end{algorithmic}
\end{algorithm}

\subsection{Permutation Property Analysis}
\label{sec:theo}
Graphs are invariant to random node permutations, and GGFlow preserves this property. To ensure permutation invariance, we analyze the permutation properties of our neural network, training objectives, and conditional probabilities path. First, we analyze the permutation invariance of the training objectives \citep{vignac2022digress}. Since the source and target distributions, along with the Hamming distance, are permutation invariant, the optimal transport map derived from Equation \ref{eq:ot} and independent coupling also exhibit this invariance.
\begin{theorem}
    \label{the:ot}
    If the distributions \( p(G^0) \) and \( p(G^1) \) are permutation invariant and the cost function maintains this invariance, then the optimal transport map \(\phi\) also respects this property, i.e., \(\phi(G^0, G^1) = \phi(\pi G^0,\pi G^1)\), where \(\pi\) is a permutation operator.
\end{theorem}
Proof of this theorem can be found in Appendix \ref{app:t1}. Thus, the training objective is permutation invariant. To ensure that the generated graph retains its identity under random permutations, the generated distribution must remain exchangeable, and GraphEvo must be permutation equivariant.
\begin{proposition}
    \label{the:propo}
    The distribution generated by the conditional flow is exchangeable with respect to nodes and graphs, i.e. $p(\bf{V}, \bf{E}) = p(\pi^* \bf{V}, \pi^* \bf{E}\pi)$, where $\pi$ is a permutation operator.
\end{proposition}

\begin{proposition}
    \label{the:graphevo}
    GraphEvo is permutation equivariant.
\end{proposition}
The proofs of Proposition \ref{the:propo} and \ref{the:graphevo} are provided in Appendix \ref{app:p2} and Appendix \ref{app:proofgraphevo}, respectively. 

\subsection{Goal-Guided Framework For Conditional Molecule Generation}
\label{sec:rl}
We propose a goal-guided framework for discrete flow matching, employing reinforcement learning (RL) to guide graph flow matching models for non-differentiable objectives. The goal of the guidance method is to map the noise distribution $p_0$ to a preference data distribution $p_1^*$ using a reward function $\mathcal{R}(G^t, t)$. 

We formulate the inference process of flow matching as a Markov Decision Process (MDP), where $(G^t, t)$ and $G^{t+\Delta t}$ are the state space $\mathbf{s}_t$ and action space $\mathbf{a}_t$, $p_0$ is an initial noise distribution, $p(G^{t+\Delta t}|G^{t}, t)$ is the transition dynamics and policy network $\pi(\mathbf{a}_t|\mathbf{s}_t)$, $\mathcal{R}(G^t, t) = r(G^{1}) \mathbb{I}[t=1]$ is the reward function

To enable exploration, we introduce a temperature parameter $T$ for the policy network during sampling, allowing the model to explore a broader space at higher temperatures:
\begin{equation}
\label{eq:RL_explore}
\begin{aligned}
    \pi(\mathbf{a}_t|\mathbf{s}_t) = \pi(G^{t+\Delta t}|G^t,t) & = \mathrm{Cat}\Big((\delta\{\cdot,G^t\} + {u}_t(\cdot,G^t|G^0,\hat{G}^1)\Delta t)/T\Big)
\end{aligned}
\end{equation}
The goal of RL training is to maximize the reward function. To prevent overfitting to the reward preference distribution, we add a Kullback–Leibler (KL) divergence term between the Reinforcement learning fine-tuned model $p_{\theta}^{RL}(\cdot)$ and pre-trained model $p_{\theta}(\cdot)$ \citep{ouyang2022training_RL_RLHF}.

We employ the policy gradient method to update the network, where the policy is refined to $\pi(\mathbf{a}_t|\mathbf{s}_t) = p_{\theta}^{(T)} ({G^{1}}|G^{t}) q(G^{t+\Delta t}|{G^{1}})$ to $\pi(\mathbf{a}_t|\mathbf{s}_t) = p_{\theta}^{(T)} ({G^{1}}|G^{t})$ \citep{sutton1999policy_RL_PG, liu2024graph}, directly increasing the probability of generating $G^1$ with higher rewards at all timestep $t$. The training objective is:
\begin{equation}
\begin{aligned}
    \mathcal{L}_{RL} = -\mathbb{E}_{p_{\theta}(G^{0:t:1})}[\alpha \mathcal{R}(G^1)\sum\limits_{t=0}^{t=1}\log p_{\theta}^{RL} (G^{1}|G^{t}) 
    - \beta \sum\limits_{t=0}^{t=1}  \mathrm{KL}(p_{\theta}^{RL}(G^{1}|G^t)||p_\theta(G^{1}|G^t))]
\label{eq:RL_update}
\end{aligned}
\end{equation}
where $p_{\theta}(G^{0:t:1})$ represents ${p_{\rm data}(G^1)\mathcal{U}(t;0,1)\pi(G^0,G^1)p_{t}(G^t|G^0,G^1)}$. Using this optimization objective, we fine-tune the pre-trained flow matching model to generate data following the preference distribution. By integrating optimal transport, we optimize the pairing of prior data and high-reward training data \citep{chen2020sequence}. The pseudo-code for the guided GGFlow training is provided in Algorithm \ref{alg:RL} and a toy example is shown in Appendix \ref{app:RL}.

\section{Experiment}
To validate the performance of our method, we compare GGFlow with state-of-the-art graph generative baselines on molecule generation and generic graph generation, over several benchmarks in Section \ref{sec:mol} and Section \ref{sec:gen}, respectively. The ability of GGFlow to perform conditional molecule generation is analyzed in Section \ref{sec:con}.

\subsection{Moleuclar Graph Generation}
\label{sec:mol}
We evaluated GGFlow on two standard molecular datasets, QM9 \citep{ramakrishnan2014quantum} and ZINC250k \citep{irwin2012zinc}, using several metrics: Validity, Validity without correction, Neighborhood Subgraph Pairwise Distance Kernel (NSPDK) Maximum Mean Discrepancy (MMD), and Frechet ChemNet Distance (FCD). To calculate these metrics, we sampled 10,000 molecules. We compared GGFlow against various molecule generation models, including GraphAF, GraphDF, MolFlow \citep{zang2020moflow}, EDP-GNN, GraphEBM \citep{liu2021graphebm}, GDSS, PS-VAE \citep{kong2022molecule}, MolHF \citep{zhu2023molhf},  GruM, SwinGNN, DiGress, and GSDM. Detailed descriptions of the datasets, baselines and metrics are provided in Appendix \ref{app:details}.

The results, presented in Table \ref{tab:mol}, indicate that GGFlow effectively captures the distribution of molecular data, showing significant improvements over the baselines. The high Validity without correction suggests that GGFlow successfully learns chemical valency rules. Additionally, GGFlow achieves superior NSPDK and FCD scores on both datasets, demonstrating its ability to generate molecules with distributions closely resembling those of natural molecules. Visualizations of molecules generated by different models are shown in Figure \ref{fig:diff_vi}, with additional results on GGFlow provided in Appendix \ref{app:visualization}.

\begin{table}[!ht]
    \small
  \caption{Generation results on the QM9 and ZINC250k datasets. Results are the means of 3 different runs. The best results and the second-best results are marked \textbf{bold} and \underline{bold}.}
  \label{tab:mol}
  \centering
  \begin{tabular}{p{1.8cm}p{0.4cm}<{\centering}p{1.2cm}<{\centering}p{0.8cm}<{\centering}cp{0.4cm}<{\centering}p{1.2cm}<{\centering}p{0.8cm}<{\centering}cp{0.9cm}<{\centering}}
    \toprule
    \multirow{2}{*}{\textbf{Method}}  & \multicolumn{4}{c}{\textbf{QM9}}  & \multicolumn{4}{c}{\textbf{ZINC250k}} & \multirow{2}{*}{\textbf{Step}}            \\
    \cmidrule(r){2-5} \cmidrule(r){6-9}
    & Val. & Val. w/o corr. & NSPDK & FCD  & Val. & Val. w/o corr. & NSPDK & FCD \\
    \midrule
    Training Set & 100 & 100 & 0.0001 & 0.040 & 100 & 100 & 0.0001 & 0.062 & -\\
    \midrule
    GraphAF & 100 & 67.14 & 0.0218 & 5.246 & 100 & 67.92 & 0.0432 & 16.128 & -   \\
    GraphDF & 100 & 83.14 & 0.0647 & 10.451 & 100 & 89.72 & 0.1737 & 33.899 & - \\
    MolFlow & 100 & 92.03 & 0.0169 & 4.536  & 100 & 63.76 & 0.0468 & 20.875 & - \\
    GraphEBM & 100 & 8.78 & 0.0287 & 6.402 & 100 & 5.29 & 0.2089 & 35.467 & - \\
    PS-VAE & - & - & 0.0077 & 1.259 & 100 & - & 0.0112 & 6.320 & - \\
    MolHF & - & - & - & - & 100 & 93.62 & 0.0387 & 23.940 & - \\
    EDP-GNN & 100 & 47.69 & 0.0052 & 2.683 & 100 & 83.16 & 0.0483 & 16.819 & 1000      \\
    GDSS & 100 & 96.17 & 0.0033 & 2.565 & 100 & {97.12} & 0.0192 & 14.032 & 1000     \\
    GSDM & 100 &  \underline{99.90} & 0.0034 & 2.614 & 100 & 92.57 & 0.0168 & 12.435 & 1000     \\
    GruM &  100 & 99.69 & \textbf{0.0002} & \underline{0.108} & 100 & \underline{98.32} & 0.0023 & \underline{2.235} & 1000 \\ 
    SwinGNN & 100 & 99.66 & \underline{0.0003} & 0.118 & 100 & 86.16 & 0.0047 & 4.398 & 500   \\
    DiGress & 100 & 98.29 & \underline{0.0003} & \textbf{0.095} & 100 & 94.98 & \underline{0.0021} & {3.482} & 500     \\
    \midrule
    GGFlow & 100 & \textbf{99.91} & \textbf{0.0002} & {0.148} & 100 & \textbf{99.63} & \textbf{0.0010} & \textbf{1.455} & 500     \\
    \bottomrule
  \end{tabular}
\end{table}

\begin{figure}[!htp]
    \centering
    \includegraphics[scale=0.24]{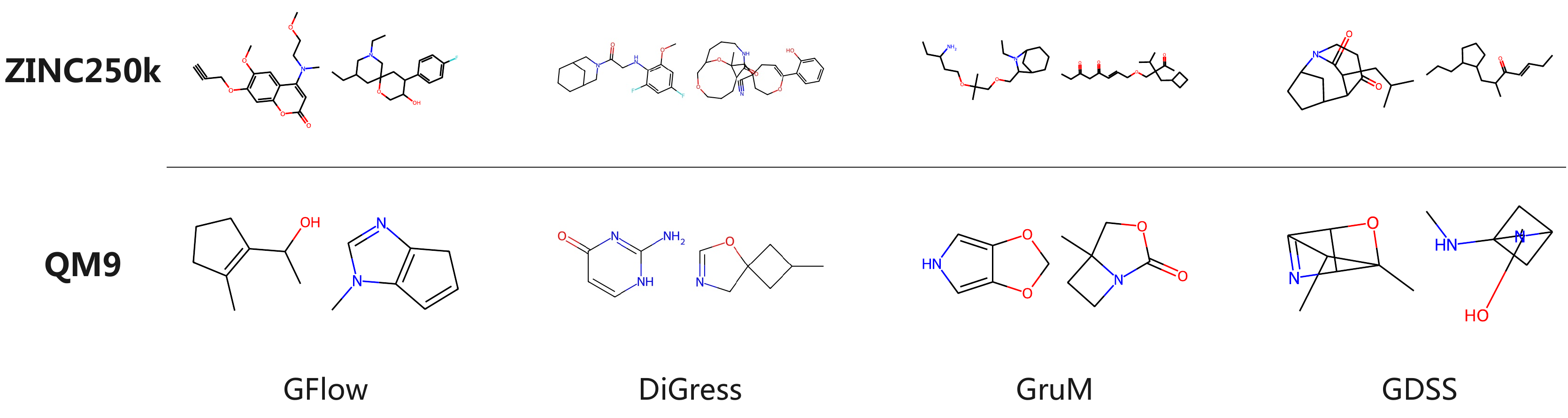}
    \caption{Visualization of generated samples of different models in different molecular datasets}
    \label{fig:diff_vi}
\end{figure}

\subsection{Generic Graph Generation}
\label{sec:gen}
We further evaluated GGFlow on three generic graph generation benchmarks of varying sizes: Ego-small, Community-small, Grid and Planar. We employ the same train/test split as GraphRNN \citep{you2018graphrnn}, utilizing 80\% of each dataset for training and the remaining for testing. We compared GGFlow's performance against well-known autoregressive models: DeepGMG \citep{li2018learning}, GraphRNN \citep{you2018graphrnn}, GraphAF \citep{shi2019graphaf}, and GraphDF \citep{luo2021graphdf} and one-shot models: GraphVAE \citep{simonovsky2018graphvae}, GNF \citep{liu2019graph}, EDP-GNN \citep{niu2020permutation}, GDSS \citep{jo2022GDSS}, DiGress \citep{vignac2022digress}, GRASP \citep{minello2024graph}, GSDM \citep{luo2023fast}, GruM \citep{jograph}, and SwinGNN \citep{yan2023swingnn}. Consistent with previous studies, we generated an equal number of graphs as the test set to compare distributions of graph statistics, including degree distribution (Deg.), clustering coefficient (Clus.), and the frequency of 4 node orbits (Orbit). Detailed descriptions of datasets, baselines, and metrics are provided in Appendix \ref{app:details}.

Table \ref{tab:gen} presents our results, showing that GGFlow achieves superior performance across most metrics. Additionally, GGFlow demonstrates comparable performance compared to state-of-the-art models in generating large graphs on the Grid dataset. These findings underscore the effectiveness of GGFlow at capturing the local characteristics and data distributions of graphs. We visualize the generated graphs in Appendix \ref{app:visualization}.

\begin{table}[!ht]
  \caption{Generation results on the generic graph datasets. Results are the means of 3 different runs. The best results and the second-best results are marked \textbf{bold} and \underline{bold}.}
  \label{tab:gen}
  \resizebox{\linewidth}{!}{
    \begin{tabular}{lccccccccccccc}
    \toprule
    \multirow{2}{*}{\textbf{Method}}  & \multicolumn{4}{c}{\textbf{Ego-small}}  & \multicolumn{4}{c}{\textbf{Community-small}} & \multicolumn{4}{c}{\textbf{Grid}} & \multirow{2}{*}{\textbf{Step}} \\
    \cmidrule(r){2-5} \cmidrule(r){6-9} \cmidrule(r){10-13}
    & Deg. & Clus. & Orbit & Avg. & Deg. & Clus. & Orbit & Avg. & Deg. & Clus. & Orbit & Avg.  \\
    \midrule
    Training Set & 0.014 & 0.022 & 0.004 & 0.013 & 0.003 & 0.009 & 0.001 & 0.005 & 0.000 & 0.000 & 0.000 & 0.000 & -\\
    \midrule
    DeepGMG & 0.040 & 0.100 & 0.020 & 0.053 & 0.220 & 0.950 & 0.400 & 0.523 & - & - & - & - & -   \\
    GraphRNN & 0.090 & 0.220 & \underline{0.003} & 0.104 & 0.080 & 0.120 & 0.040 & 0.080 & 0.064 & 0.043 & {0.021} & 0.043 & -     \\
    GraphAF & 0.031 & 0.107 & \textbf{0.001} & 0.046 & 0.178 & 0.204 & 0.022 & 0.135 & - & - & - & - & -  \\
    GraphDF & 0.039 & 0.128 & 0.012 & 0.046 & 0.060 & 0.116 & 0.030 & 0.069 & - & - & - & - & -    \\
    GNF & 0.030 & 0.100 & \textbf{0.001} & 0.044 & 0.200 & 0.200 & 0.110 & 0.170 & - & - & - & - & -     \\
    GraphVAE & 0.137 & 0.166 & 0.051 & 0.118 & 0.358 & 0.969 & 0.551 & 0.626 & 1.594 & \textbf{0.000} & 0.904 & 0.833 & - \\
    EDP-GNN & 0.054 & 0.092 & 0.007 & 0.051 & 0.050 & 0.159 & 0.027 & 0.079 & 0.460 & 0.243 & 0.316 & 0.340 & 1000 \\
    GDSS & 0.027 & \textbf{0.033} & 0.008 & \underline{0.022} & 0.044 & 0.098 & 0.009 & 0.058 & 0.133 & \underline{0.009} & 0.123 & 0.088 & 1000    \\
    GSDM & - & - & - & - & {0.020} & {0.050} & \underline{0.005} & 0.053 & \underline{0.002} &  \textbf{0.000} & \textbf{0.000} & \underline{0.001} & 1000     \\
    DiGress & 0.028 & \underline{0.046} & 0.008 & 0.027 & 0.032 & \underline{0.047} & 0.009 & \underline{0.025} & 0.037 & 0.046 & 0.069 & 0.051 & 500     \\
    SwinGNN & \underline{0.017} & 0.060 & \underline{0.003} & 0.027 & \textbf{0.006} & 0.125 & 0.018 & 0.050 & \textbf{0.000} & \textbf{0.000} & \textbf{0.000} & \textbf{0.000}  & 500  \\
    \midrule
    GGFlow & \textbf{0.005} & \textbf{0.033} & {0.004} & \textbf{0.014} & \underline{0.011} & \textbf{0.030} & \textbf{0.002} & \textbf{0.014} & {0.030} & \textbf{0.000} & \underline{0.016} & 0.015 & 500      \\
    \bottomrule
  \end{tabular}
      }
\end{table}

\subsection{Conditional Molecule Generation}
\label{sec:con}
To further evaluate the performance of our model, we conducted conditional generation experiments on the QM9 dataset, focusing on generating molecules with molecular properties \(\mu\) that closely match a target value $\mu^*$. In the experiment, we set the target value as 1, i.e. \(\mu^*=1\).

For the experiment, we employed a reinforcement learning-based guidance method and compared it to the guided version of DiGress, which also proposes an effective approach for discrete diffusion models in conditional generation tasks. The reward function was defined as \(|\mu-\mu^*|\), and the model was trained over 10,000 steps using the training settings detailed in Section \ref{sec:mol}. To evaluate the effectiveness of our guidance method, we compared it against three baselines: (1) Guidance for DiGress \citep{vignac2022digress}. (2) Direct supervised training (ST) (3) Supervised fine-tuning (SFT). Additionally, we calculated the mean and variance of  \(|\mu-\mu^*|\)for samples generated unconditionally by both DiGress and GGFlow to provide a baseline comparison. Further details of the experiment are provided in Appendix \ref{app:congen}.

The results, detailed in Table \ref{tab:con}, demonstrate the superiority of our reinforcement learning-based conditional generation method over both ST and SFT approaches. Notably, our method surpasses the guidance techniques used in diffusion models, showcasing its enhanced ability to steer the generative process toward desired outcomes. Additionally, our approach achieves higher validity in conditional generated tasks, highlighting its robustness and superior performance in goal-directed generation.

\begin{table}[!ht]
  \caption{Mean absolute error of molecular property $\mu$ on conditional generation on the QM9 dataset.}
  \label{tab:con}
  \centering
  \begin{tabular}{p{1.2cm}cccccc}
    \toprule
    \multirow{2}{*}{{Methods}} & \multicolumn{2}{c}{{DiGress}} & \multicolumn{4}{c}{{GGFlow}}   \\
    \cmidrule(r){2-3} \cmidrule(r){4-7}
    & Uncondition & +Guidance & Unconditition & Supervised Training & +SFT & +RL \\
    \midrule
    Mean & 1.562 & 1.092 & 1.569 & 1.184 & 1.223 & \textbf{0.672} \\
    Variance & 1.641 & 0.894 & 1.987 & 1.579 & 1.893 & \textbf{0.647} \\
    Val. w/o & 98.29 & 74.2 & 99.91 &  86.1 & 87.0 & \textbf{92.2} \\
    \bottomrule
  \end{tabular}
\end{table}

\section{Conclusion}
In this paper, we introduced GGFlow, a discrete flow matching generative model for molecular graphs that incorporates optimal transport and an innovative graph transformer network. GGFlow achieves state-of-the-art performance in unconditional molecular graph generation tasks. Additionally, we presented a novel guidance method using reinforcement learning to control the generative trajectory toward a preferred distribution. Furthermore, our model demonstrates the ability to achieve the best performance across various tasks with fewer inference steps compared to other baselines which highlights the practical impact of our guidance method. 
A primary limitation is scalability to larger graphs like protein ($|\mathcal{V}|>500$), attributable to the increased time complexity from triangle attention updates and spectral feature computations. Future work will focus on enhancing our model's scalability in larger graphs.


\section*{Acknowledge}
This work was supported by the National Natural Science Foundation of China (62072435, 82130055, 32271297, 32370657), the National Key Research and Development Program of China (2020YFA0907000), and Chongqing Municipal Key Special Project for Technological Innovation and Application Development (CSTB2022TIAD-DEX0035). The data underlying this article will be shared on reasonable request to the corresponding author.

\bibliographystyle{plainnat}

\newpage
\appendix
\renewcommand{\appendixname}{Appendix~\Alph{section}}
\renewcommand\thefigure{S\arabic{figure}}    
\renewcommand\thetable{S\arabic{table}}    
\section*{Appendix}
\setcounter{table}{0}   
\setcounter{figure}{0}
\section{Background}
\label{app:back}
\subsection{Continuous Flow Matching Generative Model}
\label{app:flow}

The generative model aims to establish a mapping \( f: \mathbb{R}^d \rightarrow \mathbb{R}^d \) that transforms a noise distribution \( q_0 \) into a target data distribution \( q_1 \). This transformation is dependent on a density function \( p_0 \) over \( \mathbb{R}^d \), and an integration map \( \psi_t \), which induces a pushforward transformation \( p_t = [\psi_t]_{\#}(p_0) \). This denotes the density of points \( x \sim p_0 \) transported from time 0 to time \( t \) along a vector field \( u : [0,1] \times \mathbb{R}^d \rightarrow \mathbb{R}^d \).

The vector field \( u \) is formulated as:
\begin{equation*}
\mathrm{d}x = u_t(x) \mathrm{d}t.
\end{equation*}
The solution \( \psi_t(x) \) to this ODE, with the initial condition \( \psi_0(x) = x \), represents the trajectory of the point \( x \) governed by \( u \) from time 0 to time \( t \).

The evolution of the density \( p_t \), viewed as a function \( p : [0,1] \times \mathbb{R}^d \rightarrow \mathbb{R} \), is encapsulated by the continuity equation:
\begin{equation*}
    \frac{\partial p}{\partial t} = -\nabla \cdot (p_t u_t),
\end{equation*}
with the initial condition given by \( p_0 \). Here, \( u \) is the probability flow ODE for the path of marginal probabilities \( p \), generated over time. 

In practical applications, if the probability path \( p_t(x) \) and the generating vector field \( u_t(x) \) are known and \( p_t(x) \) is tractably sampled, we leverage a time-dependent neural network \( v_\theta(\cdot, \cdot) : [0,1] \times \mathbb{R}^d \rightarrow \mathbb{R}^d \) to approximate \( u \). The neural network is trained using the flow matching objective:
\begin{equation}
    \mathcal{L}_{\rm FM}(\theta) = \mathbb{E}_{t \sim \mathcal{U}(0,1), x \sim p_t(x)} \| v_\theta(t,x) - u_t(x) \|^2,
\end{equation}
which enhances the model's capability to simulate the target dynamics accurately. Avoiding the explicit construction of the intractable vector field, recent works express the probability path as a marginal over a joint involving a latent variable $z$: $p(x_t) = \int p(z)p_{t|z}(x_t|z)$. \citep{lipman2022flow, tong2023improving} and the $p_{t|z}(x_t|z)$ is a conditional probability path, satisfying some boundary conditions at $t=0$ and $t=1$.

The conditional probability path also satisfies the transport equation with the conditional vector field $u_t(x|x_1)$:
\begin{equation}
    \frac{\partial p_t(x|x_t)}{\partial t} = - \nabla \cdot (u_t(x|x_1)p_{t}(x_t|x_1)).
\end{equation}
We can construct the marginal vector field $u_t(x)$ via the conditional probability path $p_{t|1}(x_t|x_1)$ as:
\begin{equation}
    u_t(x) = \mathbb{E}_{x_1\sim p_{1|t}}[u_t(x|x_1)].
\end{equation}
We can replace the flow matching loss $\mathcal{L}_{\rm FM}$ with an equivalent loss regressing the conditional vector field $u_t(x|x_1)$ and marginalizing $x_1$ instead: 
\begin{align*}
    \mathcal{L}_{\rm CFM}(\theta) &= \mathbb{E}_{\mathcal{U}(t;0,1),x_1\sim q, x_t\sim p_{t}(x|x_1)}[u_\theta(t,x) - u_t(x|x_1)]. \\
    \nabla_\theta \mathcal{L}_{\rm FM}(\theta) &= \nabla_\theta \mathcal{L}_{\rm CFM}(\theta). 
\end{align*}
So we can use $\mathcal{L}_{\rm CFM}(\theta)$ instead to train the parametric vector field $u_\theta$.

\section{Proofs}
\label{app:pro}
\subsection{Optimal Prior Distribution}
\label{app:prior}
This prior is structured as a product of a single distribution \( v \) for all nodes and a single distribution \( e \) for all edges, \(\prod_i v \times \prod_{i,j} e\), to ensure exchangeability across the graph components.

\begin{theorem}[Optimal prior distribution]
    \label{the:prior}
    Consider the class \( \mathcal{C} = \{\prod_i u \times \prod_{i,j} v, (u, v) \in \mathcal{P}(\mathcal{V}) \times \mathcal{P}(\mathcal{E})\} \) of distributions over graphs, which factorize as the product of a uniform distribution \( v \) over node attribute space \( \mathcal{V} \) and a uniform distribution \( e \) over edge attribute space \( \mathcal{E} \). Given any arbitrary distribution \( P \) over graphs (viewed as a tensor of order \( n + n^2 \)), with \( q_V \) and \( q_E \) as its marginal distributions for node and edge attributes respectively, then the orthogonal projection of \( P \) onto \( \mathcal{C} \) is defined as \( \phi^G = \prod_i q_V \times \prod_{i,j} q_E \). This projection minimizes the Euclidean distance:
    \begin{equation*}
        \phi^G \in \arg \min_{(v,e) \in \mathcal{C}} \| P - \prod_{1 \leq i \leq n} v \times \prod_{1 \leq i,j \leq n} e \|_2^2.
    \end{equation*}
\end{theorem}

The details and proof of Theorem \ref{the:prior} are extensively discussed in DiGress \citep{vignac2022digress}.

\subsection{Choice of conditional velocity field}
\label{app:p1}
In GGFlow, the conditional vector field for discrete flow matching is defined as \citep{campbell2024generative}:
\begin{align*}
    u_t(G, G^t|G^0,G^1) &= \frac{\text{ReLU}(\partial_t p_{t|1}(G|G^1) - \partial_t p_{t|1}(G^t|G^1))}{\mathbf{Z}_t \cdot p_{t|1}(G^t|G^1)} \\
    &= \frac{1}{\mathbf{Z}_t(1-t)p_{\rm ref}}\delta\{G,G^1\}(1-\delta\{G^t,G^1\}), G_t \neq G,
\end{align*}
where ${\rm ReLU(a) = max(a,0)}$ and \( \mathbf{Z}_t = |\{G^t: p_{t}(G^t|G^0,G^1)>0\}| \). $u_t(G, G^t|G^0,G^1)=0$ when $p_{t}(G|G^1,G^0) = 0$ and $p_{t}(G^t|G^1,G^0) = 0$. When \( G^t = G \), the rate matrix \( R(G^t, G^t | G^0,G^1) = -\sum_{G^t \neq G} R(G^t, G | G^0, G^1) \). For simplification, the graph $G$ is denoted as variable $x$
\begin{proof}
    Consider the conditional probability \( p_{t|1}(x^t|x^1,x^0) = p_{t}(x^t|x^1,x^0)={\rm Cat} \Big (t\delta\{x^1,x^t\} + (1-t)q_x \Big) \), where $q_x$ is the prior distribution. We derive its time derivative:
    \begin{equation}
        \partial_t p_{t|1}(x^t|x^1,x^0) = \delta\{x^1,x^t\} - q_x,
    \end{equation}

    We then construct the conditional rate matrix \( u_t(x^t, x|x^1,x^0) \) as:
    \begin{align*}
        u_t(x^t, x|x^1,x^0) &= \frac{\text{ReLU}(\partial_t p_{t|1}(x|x^1,x^0) - \partial_t p_{t|1}(x^t|x^1,x^0))}{\mathbf{Z}_t \cdot p_{t|1}(x^t|x^1,x^0)} \\
        &= \frac{\text{ReLU}(\delta\{x,x^1\} - q_x - \delta\{x^t,x^1\} + q_x)}{\mathbf{Z}_t (t\delta\{x^1,x^t\} + (1-t)q_x)} \\
        &= \frac{\text{ReLU}(\delta\{x,x^1\} - \delta\{x^t,x^1\})}{\mathbf{Z}_t (t\delta\{x^1,x^t\} + (1-t)q_x)}.
    \end{align*}

    The expression simplifies under the assumption that \( x^t \neq x \). The only non-zero values occur when \( x = x^1 \) and \( x^t \neq x^1 \), thus yielding:
    \begin{equation}
        u_t(x^t, x|x^1) = \frac{1}{\mathbf{Z}_t(1-t)q_x}\delta\{x,x^1\}(1-\delta\{x^t,x^1\}), x_t \neq j 
    \end{equation}
    where \( \mathbf{Z}_t = |\{x^t: p_{t|1}(x^t|x^1,x^0) > 0\}| \).
\end{proof}

\subsection{Proof of Proposition \ref{the:propo}}
\label{app:p2}
\begin{proof}
    The Kolmogorov forward equations for discrete flow matching are expressed as:
    \begin{equation}
        \partial_t p_t = u_t p_t,
    \end{equation}
     If we establish the permutation invariance of the prior distributions \( p_{\rm ref} \) and the permutation equivariance of conditional flow probabilities, then it follows that \( p(G^1) \) is permutation exchangeable.

    According to the Theorem \ref{the:prior}, we deduce the permutation invariance of the prior distribution \( p_{\rm ref} \). Given the conditional probabilities \( p(G^{t+\Delta t}|G^t) = {\rm Cat}\Big (\delta\{G^t, G^{t+\Delta t}\} + \hat{u}_t(G^t, G^{t+\Delta t})\Delta t \Big ) \), it suffices to demonstrate the permutation equivariance of the conditional probabilities. This requires showing the permutation equivariance of the vector field \( u_t \). Consider the case for nodes:
    \begin{align*}
        \pi u_t^V(V_i^t, V_i^{t+\Delta t}) &= \pi\left(\mathbb{E}_{\hat{p}_{1|t}^V(V_i^1|V_i^t)}[u_t^V(V_i^t, V_i^{t+\Delta t}|V_i^1,V_i^0)]\right), \\
        \text{LHS} &= u_t^V(V_{\pi^{-1}(i)}^t, V_{\pi^{-1}(i)}^{t+\Delta t}), \\
        \text{RHS} &= \left(\mathbb{E}_{\hat{p}_{1|t}^V(V_{\pi^{-1}(i)}^1|V_{\pi^{-1}(i)}^t)}[u_t^V(V_{\pi^{-1}(i)}^t, V_{\pi^{-1}(i)}^{t+\Delta t}|V_{\pi^{-1}(i)}^1,V_{\pi^{-1}(i)}^0)]\right), \\
        &= u_t^V(V_{\pi^{-1}(i)}^t, V_{\pi^{-1}(i)}^{t+\Delta t}) = \text{LHS}.
    \end{align*}
    where \( \pi \) is a permutation operator. This establishes the permutation equivariance of \( u_t \) and the exchangeability of the generated distribution.
\end{proof}

\subsection{Proof of Theorem \ref{the:ot}}
\label{app:t1}
\begin{proof}
    Building on the foundations established in Theorem \ref{the:prior} and Proposition \ref{the:propo}, we confirm the permutation invariance of both the target and source distributions. The Hamming distance is invariance under random permutations \(\pi\), as shown by:

    \begin{align*}
        H(G^0, G^1) &= \sum_{i} \delta(v_i^0, v_i^1) + \frac{1}{2}\sum_{i,j} \delta(e_{ij}^0, e_{ij}^1) \\
        &= \sum_{i} \delta(v_{\pi^{-1}(i)}^0, v_{\pi^{-1}(i)}^1) + \frac{1}{2}\sum_{i,j} \delta(e_{\pi^{-1}(i)\pi^{-1}(j)}^0, e_{\pi^{-1}(i)\pi^{-1}(j)}^1) \\
        &= H(\pi G^0, \pi G^1)
    \end{align*}
    This property of the Hamming distance ensures the permutation invariance of the optimal transport map \(\phi^*\).

\end{proof}

\section{Details of GraphEvo}
\label{app:graphevo}
GraphEvo is a novel edge-augmented graph transformer model designed for graph data. To enhance the generative capabilities of GGFlow, GraphEvo introduces a triangle update mechanism, which significantly improves the exchange of edge information. We incorporate FiLM and PNA layers into our architecture \citep{vignac2022digress}:
\begin{align*}
    {\rm FiLM}(X_1,X_2) &= X_1({\rm Linear}(X_2)+1) + {\rm Linear'}(X_2) \\{\rm PNA}(X) &= {\rm Linear}\Big ({\rm Cat}(\max (X),\min (X), {\rm mean}(X), {\rm std}(X))\Big ).
\end{align*}
The full architecture of GraphEvo is illustrated in Algorithm \ref{alg:whole} and is permutation equivariant. The time complexity of GraphEvo is $O(N^3)$.
\begin{algorithm}[!htbp]
    \caption{Architecture of GraphEvo}\label{alg:whole}
    \begin{algorithmic}[1]
        \REQUIRE $G, t, N_{\rm layer}$
        \STATE $\mathbf{V}, \mathbf{E} \leftarrow G$ 
        \STATE $\mathbf{y} \leftarrow {\rm ExtractFeature}(G), \mathbf{t} \leftarrow {\rm TimeEmbedding}(t)$
        \STATE $\mathbf{y} \leftarrow \mathbf{y} + \mathbf{t}$
        \STATE $\mathbf{X}, \mathbf{E}, \mathbf{y} \leftarrow {\rm Linear}(V),{\rm Linear}(\mathbf{E}), {\rm Linear}(\mathbf{y})$
        \FOR{$t = 0, 1, \dots, N_{\rm layer}$}
        \STATE $\mathbf{X'}, \mathbf{E'}, \mathbf{y'} \leftarrow {\rm SelfAttention}(\mathbf{X}, \mathbf{E}, \mathbf{y} )$
        \STATE $\mathbf{X} \leftarrow {\rm ReLU}\Big({\rm LayerNorm}(\mathbf{X} + {\rm Dropout}(\mathbf{X'}))\Big)$
        \STATE $\mathbf{E} \leftarrow {\rm ReLU}\Big({\rm LayerNorm}(\mathbf{E} + {\rm Dropout}(\mathbf{E'}))\Big)$
        \STATE $\mathbf{y} \leftarrow {\rm ReLU}\Big({\rm LayerNorm}(\mathbf{y} + {\rm Dropout}(\mathbf{y'}))\Big)$
        \ENDFOR
        \STATE $\hat{p}_{1|t}^V(V^1|V^t,V^0), \hat{p}_{1|t}^E(E^1|E^t,E^0), \mathbf{y} \leftarrow {\rm Linear}(V),{\rm Linear}(\mathbf{E}), {\rm Linear}(\mathbf{y})$
        \RETURN $\hat{p}_{1|t}^V(V^1|V^t,V^0), \hat{p}_{1|t}^E(E^1|E^t,E^0),\mathbf{y}$
    \end{algorithmic}
\end{algorithm}

GraphEvo integrates global structural features to improve generation performance, including both graph-theoretic and domain-specific attributes:

\textbf{Graph-theoretic features}: These encompass node-level properties such as the number of $k$-cycles ($k \leq 5$) containing this point and an estimate of the largest connected component, alongside graph-level metrics like the total number of $k$-cycles ($k \leq 6$) and connected components.

\textbf{Molecular features}: These account for the current valency of each atom and the molecular weight of the entire molecule.

\subsection{Proof of Proposition \ref{the:graphevo}}
\label{app:proofgraphevo}
\begin{proof}
    Let $G^t=(V^t,E^t)$ is a intermediate graph, and $\pi G^t=(\pi^*V,\pi^*E\pi)$ is the permutation. To prove the permutation properties of the graph, we need to consider two aspects: additional structural features and the model architecture.
    
    First, the spectral and structural features are permutation equivariant for node-level features and invariant for graph-level features. Additionally, the FiLM blocks and Linear layers are permutation equivariant, while the PNA pooling function is permutation invariant. Layer normalization is also permutation equivariant.

    As GraphEvo is built using permutation equivariant components, we conclude that the overall model is permutation equivariant.

\end{proof}



\section{Toy example of goal-guided graph generation}
\label{app:RL}
\begin{figure}[!htp]
    \centering
    \includegraphics[scale=0.75]{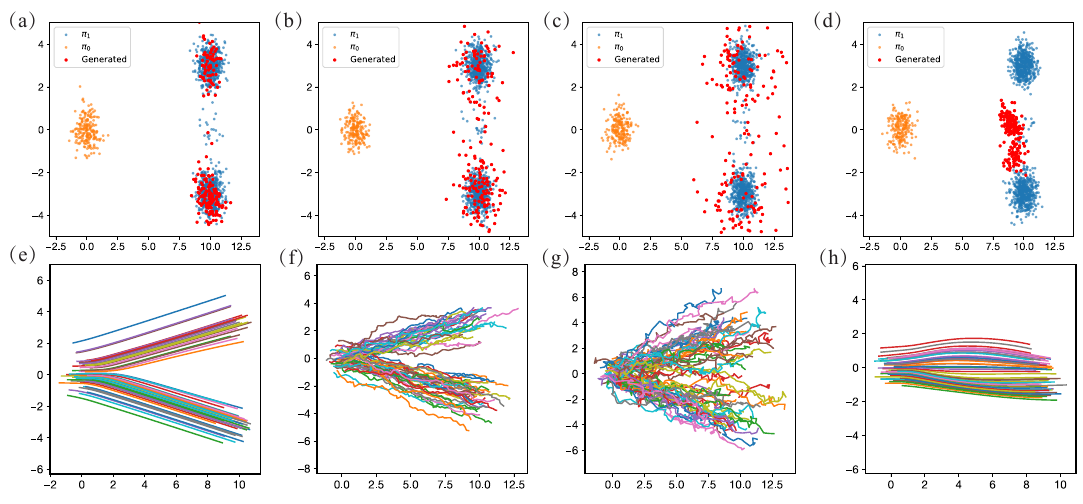}
    \caption{(a-d) Data distribution of the flow matching model, $\pi_0$ is the original distribution (orange), $\pi_1$ is the target data distribution (blue), and the red dots are the data distribution generated by the model. (e-h) In reinforcement learning, the flow matching model conducts exploration/sampling trajectories}
    \label{fig:RL_demo}
\end{figure}

We demonstrate the utility of our goal-guided framework of flow matching with a toy example, depicted in Figure \ref{fig:RL_demo}: (a) shows a trained unconditional flow matching model mapping noise distribution \(p_0 \) to data distribution \(p_1\). (b, c) illustrate the effect of temperature $T$ on the exploration, with higher temperatures resulting in broader data point distribution. (d) shows how fine-tuning according to Equation \ref{eq:RL_update} concentrates data in regions with higher rewards. (e-f) illustrate the corresponding flow matching trajectories.

\section{Implement Details}
\label{app:details}
\subsection{Algorithms of GGFlow}
Details of the training procedure and guided training procedure are provided in Algorithm \ref{alg:train} and \ref{alg:RL}.
\begin{algorithm}[h]
    \caption{Training Procedure of GGFlow}\label{alg:train}
    \begin{algorithmic}[1]
        \REQUIRE $G = (V,E), q_V, q_E,$
        \FOR {$n \in \{0,\dots,N_{\mathrm{iter}}-1\}$}
        \STATE $t \in \mathcal{U}(0,1),G^1= G$
        \STATE $G^0=(V^0,E^0) \sim p^{\rm ref}$
        \STATE $(G^0, G^1) \sim {\rm OptimalTransport}(G^0,G^1)$
        \STATE // Sample from conditional probability flow.
        \STATE $V^t = (t\delta\{V^1,\cdot\}+(1-t)V^0)$ and $E^t = (t\delta\{E^1,\cdot\}+(1-t)E^0)$ 
        \STATE $\hat{p}_{1|t}^V(V^1|V^t, V^0), \hat{p}_{1|t}^E(E^1|E^t,E^0),\mathbf{y} = {\rm GraphEvo}_{\theta_n}(V^t,E^t,t,f^t)$
        \STATE $\mathcal{L} = \mathbb{E}_{p_{\rm data}(G^1)\mathcal{U}(t;0,1)\pi(G^0,G^1)p_{t}(G^t|G^0,G^1)}[\log \hat{p}_{1|t}(G^1|G^t,G^0)]$
        \STATE $\theta_{n+1} = {\rm optimizer\_update(\theta_n,\mathcal{L})}$
        \ENDFOR
        \STATE $\theta^* = \theta_{N_{\rm iter}}$
        \RETURN $\theta^*$
    \end{algorithmic}
\end{algorithm}

    

\begin{algorithm}[h]
    \caption{Training Procedure of Guided GGFlow by Reinforcement Learning}\label{alg:RL}
    \begin{algorithmic}[1]
        \REQUIRE $\theta_0$, $\theta$, $\alpha, \beta, T$, $N_{\rm steps}$, $\mathrm{traj}$,  $G^0\sim p_{\rm ref}, u_t(G^t,G|G^1,G^0)$, $T$, $N_{\rm train}$
        \STATE $\theta \gets \theta_0$
        \FOR{$i \in \{1,\dots, N_{\rm train}\}$}
        \STATE $\Delta t = 1/N_{\rm steps}$
        \STATE Collect flow trajectory $\Big(G^{0}, t=0, \mathcal{R}(G^0)\Big)$.
        \FOR{$n \in \{0,\dots,N_{\mathrm{steps}}-1\}$}
        \STATE $\hat{p}_{1|t}^V(V^1|V^t,V^0), \hat{p}_{1|t}^E(E^1|E^t,E^0), \mathbf{y} = {\rm GraphEvo}(V^t,E^t,t)$
        \STATE Get $G^{t+\Delta t}$ by sampling from Equation \ref{eq:RL_explore}.
        \STATE $(V^{t+\Delta t}, E^{t+\Delta t}) = G^{t+\Delta t}$
        \STATE $t = t+\Delta t$
        \STATE Compute the reward function $\mathcal{R}(G^{t+\Delta t})$.
        \STATE Collect flow trajectory $\Big(G^{t+\Delta t}, t+\Delta t, \mathcal{R}(G^{t+\Delta t})\Big)$.
        \ENDFOR
        \STATE Update network using Equation \ref{eq:RL_update}.
        \STATE $t=0$
        \ENDFOR
        \RETURN Guided flow matching model $\theta^*$
    \end{algorithmic}
    
\end{algorithm}

\subsection{Baselines Implementation}
\label{app:baselines}
To benchmark the performance of GGFlow, we ensure consistency by using identical splits of training and test sets across all datasets. Below, we provide the implementation details for each baseline model. To guarantee a fair comparison, most baseline models are retrained three times, and the average results from these runs are reported as the final outcomes in unconditional generation tasks. The results of the DeepGMG, GraphRNN and GNF for Ego-small and Community-small dataset are taken from their original papers.



\textbf{GraphAF} \citep{shi2019graphaf}\quad We follow the implementation guidelines provided in the TorchDrug tutorials (\url{https://torchdrug.ai/docs/tutorials/generation.html}).

\textbf{GraphDF} \citep{shi2019graphaf}\quad Model scripts are sourced from the DiG repository (\url{https://github.com/divelab/DIG/tree/dig-stable/examples/ggraph/GraphDF}).

\textbf{GraphVAE} \citep{shi2019graphaf}\quad Scripts are obtained from the GraphVAE section of the GraphRNN repository (\url{https://github.com/JiaxuanYou/graph-generation/tree/master/baselines/graphvae}).


\textbf{MoFlow} \citep{zang2020moflow}\quad Implementation scripts are taken from the MoFlow repository (\url{https://github.com/calvin-zcx/moflow}).

\textbf{GraphEBM} \citep{liu2021graphebm}\quad We use the implementation available in the GraphEBM repository (\url{https://github.com/biomed-AI/GraphEBM}).

\textbf{EDP-GNN} \citep{niu2020permutation}\quad The model is implemented according to the scripts in the EDP-GNN repository (\url{https://github.com/ermongroup/GraphScoreMatching}).

\textbf{GDSS} \citep{jo2022score}\quad Implementation details are sourced from the GDSS repository (\url{https://github.com/harryjo97/GDSS}).

\textbf{GSDM} \citep{luo2023fast}\quad Scripts are implemented from the GSDM repository (\url{https://github.com/ltz0120/Fast_Graph_Generation_via_Spectral_Diffusion}).

\textbf{PS-VAE} \citep{kong2022molecule} \quad Implementation details are sourced from the PS-VAE repository (\url{https://github.com/THUNLP-MT/PS-VAE}).

\textbf{MolHF} \citep{zhu2023molhf} \quad The model is implemented according to the scripts in the MolHF repository (\url{https://github.com/violet-sto/MolHF}).

\textbf{GRASP} \citep{minello2024graph} \quad Implementation details are sourced from the GRASP repository (\url{https://github.com/lcosmo/GRASP}).

\textbf{SwinGNN} \citep{yan2023swingnn} \quad Implementation details are sourced from the SwinGNN repository (\url{https://github.com/DSL-Lab/SwinGNN}). The authors employ the 'gaussian\_tv' MMD kernel, whereas other methods use 'gaussian\_emd' or 'gaussian'. To ensure a fair comparison, we adopt the same kernel.

\textbf{GruM} \citep{jograph} \quad Scripts are implemented from the GruM repository (\url{https://github.com/harryjo97/GruM/}).

\textbf{DiGress} \citep{vignac2022digress}\quad The implementation is based on the DiGress repository (\url{https://github.com/cvignac/DiGress}).

\subsection{Details of Molecule Datasets}
\label{app:molecule_datasets}
\subsubsection{Dataset}
\textbf{QM9}\quad It is a subset of the GDB-17 database and consists of 134,000 stable organic molecules, each containing up to 9 heavy atoms: carbon, oxygen, nitrogen, and fluorine \citep{ramakrishnan2014quantum}. The dataset includes 12 tasks related to quantum properties. We follow the train/test split from GDSS, using 12,000 molecules for training and the remaining 1,000 for testing.

\textbf{ZINC250k}\quad It contains 250,000 drug-like molecules with a maximum of 38 atoms per molecule \citep{irwin2012zinc}. It includes 9 atom types and 3 edge types. For a fair comparison, we use the same train/test split as previous works, such as GDSS and GSDM.

\begin{table}[!ht]
  \caption{Statistics of the molecular graph datasets}
  \label{tab:sta}
  \centering
  \resizebox{\linewidth}{!}{
  \begin{tabular}{lccccc}
    \toprule
    Dataset  & type & Number of graphs & Number of nodes & Number of node types & Number of edge types\\
    \midrule
    QM9 & Real& 133,885 & [1,  9] & 4 & 3  \\
    ZINC250k & Real & 249,455 & [6, 38] & 9 & 3  \\
    \bottomrule
  \end{tabular}}
\end{table}

\subsection{Details of Generic Datasets}
\label{app:gen}
\subsubsection{Dataset}

\textbf{Ego-small}\quad This dataset consists of 200 small one-hop ego graphs derived from the Citeseer network \citep{sen2008collective}. Each graph contains between 4 and 18 nodes.

\textbf{Community-small}\quad This dataset includes 100 random community graphs, each formed by two communities of equal size generated using the E-R model \citep{erdHos1960evolution} with a probability parameter of $p=0.7$. The graphs range in size from 12 to 20 nodes.


\textbf{Grid}\quad The dataset consists of 100 standard 2D grid graphs with $100 \leq |V | \leq 400$.


\begin{table}[!ht]
  \caption{Statistics of the generic graph datasets}
  \label{tab:sta_gen}
  \centering
  \begin{tabular}{lccc}
    \toprule
    Dataset  & type & Number of graphs & Number of nodes \\
    \midrule
    Ego-small & Real & 200 & [4, 18]  \\
    Community-small & Synthetic & 100 & [12, 20]  \\
    Grid & Synthetic & 100 & [100,400] \\
    
    \bottomrule
  \end{tabular}
\end{table}

\subsubsection{Metrics}
\textbf{Validity}\quad We permit atoms to exhibit formal charges during valency checks because of the presence of formal charges in the training molecules. It is the fraction of valid molecules after valency correction or edge resampling. 

\textbf{Validity w/o correction}\quad This metric explicitly evaluates the quality of molecule generation before any correction phase, providing a baseline for raw generation performance.

\textbf{FCD}\quad FCD quantifies the functional connectivity density within a molecule by computing distances and connectivity between atoms, based on both structural and chemical features. It describes the three-dimensional structure, topological features, and chemical properties of molecules, making it valuable in fields such as drug design, compound screening, and molecular simulations.

\textbf{NSPDK}\quad  NSPDK assesses molecular similarity by comparing shortest paths within their graphical structures. It captures connectivity patterns and chemical environments, effectively describing relationships and similarities between molecules. For two distributions $p$ and $q$, the MMD using NSPDK is calculated as:

\begin{align} 
\text{MMD}^2_{\text{NSPDK}}(p, q) =& \frac{1}{n(n-1)} \sum_{i=1}^{n} \sum_{j \neq i}^{n} k_{\text{NSPDK}}(\mathcal{X}_i, \mathcal{X}_j) + \frac{1}{m(m-1)} \sum_{i=1}^{m} \sum_{j \neq i}^{m} k_{\text{NSPDK}}(\mathcal{Y}_i, \mathcal{Y}_j) \\
 &- \frac{2}{mn} \sum_{i=1}^{n} \sum_{j=1}^{m} k_{\text{NSPDK}}(\mathcal{X}_i, \mathcal{Y}_j) 
\end{align}

Here, \( k_{\text{NSPDK}}(\cdot) \) denotes the NSPDK kernel function.  \( \mathcal{X} \) is the set of molecules from distribution \( p \). \( \mathcal{Y} \) is the set of molecules from distribution \( q \). \( n \) and \( m \) represent the number of samples drawn from distributions \( p \) and \( q \), respectively. This formula quantifies the difference between the distributions \( p \) and \( q \) using the NSPDK kernel.

For generic graph datasets, we employ Maximum Mean Discrepancy (MMD) to assess the distributions of graph statistics, specifically degree distribution, clustering coefficient, the number of occurrences of 4-node orbits, and eigenvalues of the normalized graph Laplacian. In alignment with prior research \citep{jo2022score}, we utilize specialized kernels for MMD calculations: the Gaussian Earth Mover's Distance (EMD) kernel for degree distribution and clustering coefficient, the Gaussian Total Variation (TV) kernel for eigenvalues of the normalized graph Laplacian, and a standard Gaussian kernel for the 4-node orbits. To ensure a fair comparison, the size of the prediction set matches that of the test set.

\subsection{Details of Conditional Generation}
\label{app:congen}
We included three guidance baselines in our conditional generation task:

\textbf{DiGress model with guidance}\quad Utilizing the guidance method integrated into the DiGress model \citep{vignac2022digress}. 

\textbf{Direct supervised training (ST)}\quad It involved selecting training samples from the dataset that satisfied \(|\mu-\mu^*| < 1.0\) and retraining them using supervised learning settings identical to those in Section \ref{sec:mol}. 

\textbf{Supervised fine-tuning (SFT)}\quad This method involved fine-tuning a pre-trained GGFlow model on molecules generated with \(|\mu-\mu^*| < 1.0\), maintaining the same training settings as in Section \ref{sec:mol}.

These models were trained over 10,000 steps using the training settings detailed in Section \ref{sec:mol}. We then generated 1,000 samples to calculate the results for each guidance method and the unconditional method, with the values of $\mu$ estimated using Psi4 \citep{smith2020psi4}. We set the hyperparameters $\alpha$ and $\beta$ as 0.999 and 0.001.

\section{Visualization}
\label{app:visualization}
\begin{figure}[!htp]
    \centering
    \includegraphics[scale=0.15]{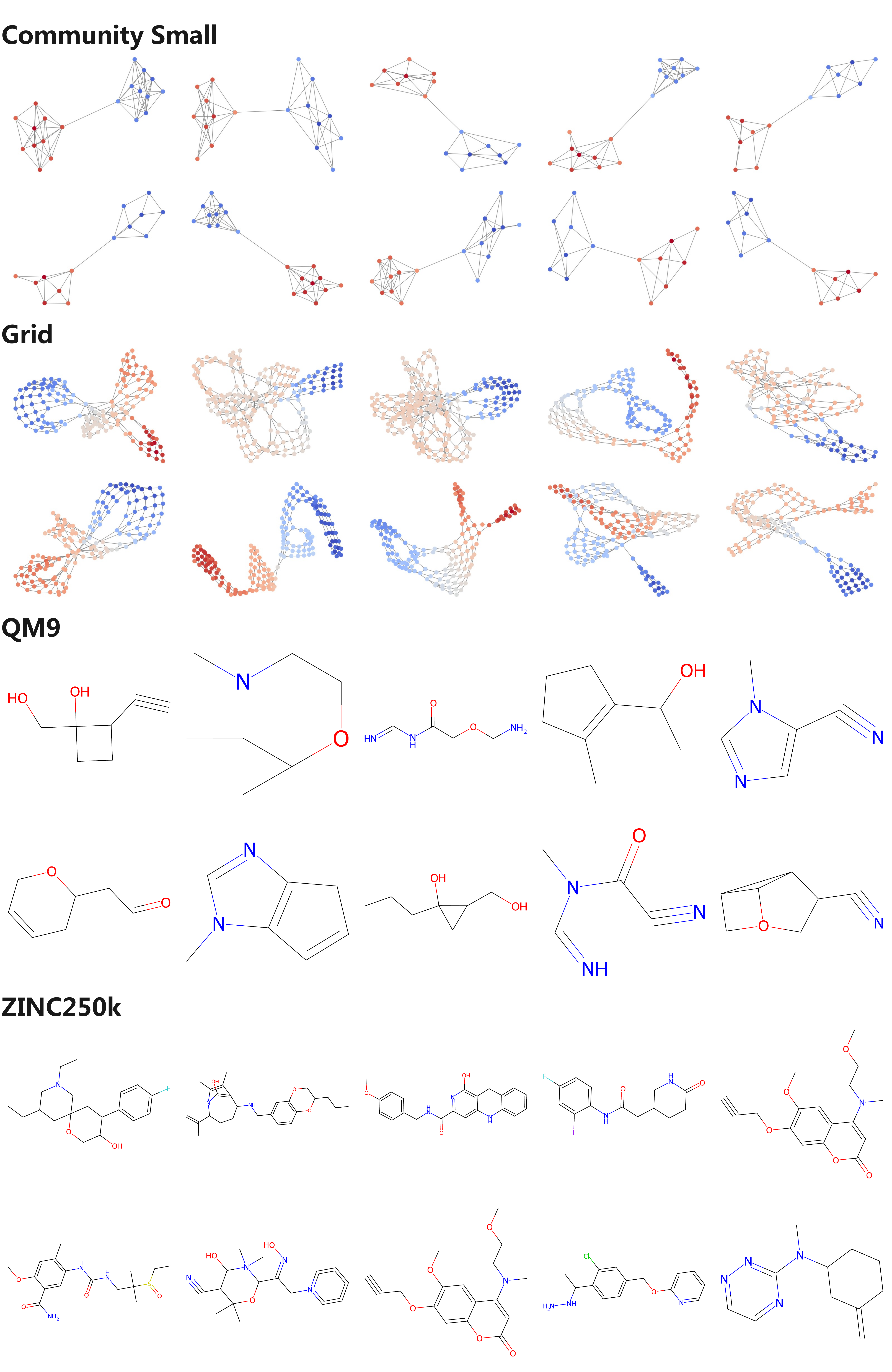}
    \caption{Visualization of generated samples of our model in different datasets}
    \label{fig:vi}
\end{figure}

\end{document}